\documentclass[onecolumn]{svjour3}

\smartqed  

\usepackage[a4paper,margin=25mm]{geometry}

\usepackage[misc]{ifsym} 
\usepackage{tabularx}
\usepackage[style=authoryear, url=false, isbn=false, doi=false, eprint=false,backend=biber, maxcitenames=1]{biblatex}
\usepackage[figuresright]{rotating}
\usepackage{todonotes}
\usepackage{longtable}
\usepackage{float}
\usepackage{caption}
\usepackage{subcaption}
\usepackage{url}
\usepackage[noend,ruled,vlined,linesnumbered]{algorithm2e}
\usepackage{amsmath}
\usepackage{hyperref}
\usepackage{multicol}
\usepackage{wrapfig}

\newcommand{\orcid}[1]{\href{https://orcid.org/#1}{\protect\includegraphics[width=8pt]{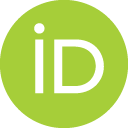}}}
\newcommand{\DWorcid}{0000-0001-8809-3587}
\newcommand{\MKorcid}{0000-0002-0997-5889}
\newcommand{\VKorcid}{0000-0001-8405-269X}
\newcommand{\LPorcid}{0000-0002-4286-3870}

\journalname{Operational Research}

\begin{document}

\title{Variable Neighborhood Search for the Electric Vehicle Routing Problem}

\author{David Woller$^{1}$\orcid{\DWorcid} \and
        Viktor Kozák$^{1,2}$\orcid{\VKorcid} \and
        Miroslav Kulich$^1$\orcid{\MKorcid} \and
        Libor Přeučil$^1$\orcid{\LPorcid}
}


\institute{
\begin{tabular}{@{}ll}
\Letter & David Woller \\
        & wolledav@cvut.cz \\
        &   \\
$^1$    & Czech Institute of Informatics, Robotics, and Cybernetics, Czech Technical University \\
& in Prague, Jugoslávských partyzánů 1580/3, Praha 6, 160 00, Czech Republic \\
$^2$    & Department of Cybernetics, Faculty of Electrical Engineering, Czech Technical \\ & University in Prague, Karlovo náměstí 13, Praha 2, 121 35, Czech Republic \\
\end{tabular}
}

\date{}

\maketitle

\begin{abstract}
The Electric Vehicle Routing Problem (EVRP) extends the classical Vehicle Routing Problem (VRP) to reflect the growing use of electric and hybrid vehicles in logistics. Due to the variety of constraints considered in the literature, comparing approaches across different problem variants remains challenging. A minimalistic variant of the EVRP, known as the Capacitated Green Vehicle Routing Problem (CGVRP), was the focus of the CEC-12 competition held during the 2020 IEEE World Congress on Computational Intelligence. This paper presents the competition-winning approach, based on the Variable Neighborhood Search (VNS) metaheuristic. The method achieves the best results on the full competition dataset and also outperforms a more recent algorithm published afterward.

\keywords{Electric Vehicle Routing Problem \and Capacitated Green Vehicle Routing Problem \and Variable Neighborhood Search \and Metaheuristics \and Combinatorial Optimization}
\end{abstract}
\noindent

\section{Introduction}\label{sec:Introduction}

In terms of tonne-kilometers, the largest portion of all goods worldwide is transported by sea (\textcite{maritime}).
However, the road transport sector is responsible for more than 11\% of all global greenhouse gas emissions, more than five times the amount produced by aviation or shipping transport (\textcite{roser_2020}).
Electric vehicles have gradually become a usable, economically viable, and more environmentally friendly alternative to conventional vehicles in some applications during the last decade.
Therefore, numerous major logistics and delivery companies are experimenting with their deployment on a large scale.
Due to their specifics, e.g., limited range and sparse recharging infrastructure, new problems emerge in operations research, such as the Electric Vehicle Routing Problem (EVRP).

The IEEE WCCI 2020 CEC-12 Competition on Electric Vehicle Routing Problem (\textcite{competition}) aims to promote research on EVRP.
The formulation selected for the competition is a seldom addressed one, yet fundamental. 
It combines CVRP and GVRP.
In the CVRP, vehicles have limited cargo capacity and may need to visit a central depot for cargo replenishment.
The GVRP assumes unlimited cargo capacity but considers limited vehicle driving range and sparse infrastructure of alternative fuel stations (AFS).
Vehicles can recharge on the go and thus increase their range without visiting the central depot.
In both cases, the goal is to minimize the total distance traveled.
The combined problem addressed in this paper is called EVRP to remain consistent with competition but is often called CGVRP in the literature.
Additionally, a publicly available benchmark dataset is provided.
The competition omits all but the most common constraints used in other EVRP formulations.
This approach goes against the trend evident from the literature - most of the authors focus on adding completely new constraints or combining many of them.
This approach enriches the portfolio of solvable EVRP variants but typically does not allow for direct comparison of various methods, as each author uses a slightly different formulation.
Thus, the competition's primary goal is to enable a simple comparison of various approaches to the EVRP rather than creating an entirely new problem.
The contributions in this paper are the following.
\begin{itemize}
    \item Adapting the VNS for the EVRP \\
    The proposed method is based on a well-established VNS metaheuristic introduced by~\textcite{MLADENOVIC19971097}.
    It combines components from methods designed for the VRP or similar problems (permutation-based local search operators, perturbation mechanism) and problem-specific components designed specifically for the EVRP (local search operators AFS-realloc-1 and AFS-realloc-more, construction and repair procedure).
    \item Proposing a novel three-step initial solution construction \\
    A robust method for constructing a high-quality valid initial solution is proposed.
    It is based on the previous work described in~\textcite{Kozak2021}, where a large number of various construction methods for VRP, Capacitated VRP (CVRP), and Capacitated Green VRP (CGVRP) were compared.
    \item Proposing a novel repair procedure \\
    A repair procedure based on the one of~\textcite{Zhang2018} is utilized in the VNS.
    However, the original procedure fails to produce a valid solution in some instances, as shown in~\textcite{Kozak2021}.
    Thus, a modified robust version is proposed in this paper.
    \item Documented design process \\
    The proposed VNS evolved from the Greedy Randomized Adaptive Search Procedure (GRASP,~\textcite{Feo1989}) described in~\textcite{Woller2021}.
    The proposed VNS share some ideas with the GRASP (e.g., local search heuristic and some standard local search operators), but there are significant differences.
    The VNS uses a more complex three-stage initial solution construction, several problem-specific local search operators, and a perturbation mechanism for escaping local optima.
    The evolution from GRASP to VNS is carried out systematically and is documented by experiments; therefore, the benefit brought about by individual components is easily distinguishable.
    \item State of the art method \\ 
    Finally - the presented VNS consistently outperforms the remaining methods submitted to the EVRP competition, the GRASP method from~\textcite{Woller2021}, and the Bilevel Ant Colony Optimization (BACO) method published after the competition (\textcite{Jia2021}).
Thus, it can be considered state of the art for this fundamental EVRP variant.
\end{itemize}

The paper is structured as follows.
Section~\ref{sec:Introduction} contains this introduction and discusses related works.
Section~\ref{sec:prob_form} provides the mathematical formulation of the problem.
Section~\ref{sec:methods} describes all components of the designed VNS metaheuristic.
Section~\ref{sec:results} documents the evolution of the VNS, presents the competition results, and additional experiments on other datasets.
Finally, Section~\ref{sec:conclusions} contains the conclusions.

\subsection{Related works}\label{sec:related_works}

The Vehicle Routing Problem (VRP) is one of the classical combinatorial optimization problems, traceable back to~\textcite{Dantzig1959}.
Due to its $\mathcal{NP}$ hardness, the number of practical applications, and the interesting variants (\textcite{Toth2014}), it is still a challenging and relevant problem even today.
The Electric Vehicle Routing Problem (EVRP) is one of the most recent variants, loosely reflecting the integration of electric vehicles (EVs) into major delivery companies' fleets.
A recent survey by~\textcite{Erdelic2019} of 79 papers on EVRP shows that the terminology is ambiguous and that a wide variety of aspects of the operation of electric vehicles are addressed in the literature.
These are, for example: 

\begin{itemize}
    \item considering the limited driving range of EVs and the recharging time (\textcite{Schneider2015});
    \item allowing partial recharging, battery swapping or hybrid vehicles (\textcite{Felipe2014});
    \item modeling fuel consumption as a function of current load, speed, or tour (\textcite{Basso2019});
    \item acknowledging the non-linearity of the recharging process or even its influence on battery lifespan (\textcite{Froger2019});
    \item minimizing the environmental impact, fuel consumption, total distance traveled, total delivery time or total operation cost (\textcite{Li2018}).
\end{itemize}

When combined with other constraints commonly used in VRP problems, such as limited fleet size, vehicle capacity, or customer time windows, it is apparent why the resulting number of newly emerged EVRP variants is so large.
Some authors are quite successful in reflecting this by gradually generalizing their models and incorporating a large portion of the constraints (such as~\textcite{Schneider2014},~\textcite{Goeke2015},~\textcite{Schneider2015} and~\textcite{Hof2017}).

Regarding solution methods, both exact methods and metaheuristics are widely applied.
For some variants of the VRP with fewer constraints, medium-sized instances are solvable to optimality by exact methods -~\textcite{Pecin2017} reported success on instances with up to 360 customers for the Capacitated VRP, whereas~\textcite{Baldacci2012} managed up to 100 customers for the VRP with Time Windows.
However, the reported numbers for some EVRP variants are less favorable since only instances with up to 30 customers are currently solvable (\textcite{Desaulniers2016},~\textcite{Hiermann2016}).
As for metaheuristics, a broad spectrum of methods was applied to some variants of the EVRP.
According to the survey by~\textcite{Erdelic2019}, the following methods were applied at least five times: Adaptive Large Neighborhood Search, Genetic Algorithm, Large Neighborhood Search, (Iterated) Tabu Search, and Variable Neighborhood Search. 
In general, neighborhood-oriented metaheuristics have been reported to outperform population-based methods.

This paper uses the EVRP formulation of the IEEE WCCI 2020 competition on the EVRP assigned by~\textcite{tech_report}.
The competition formulation omits all but the two most essential constraints of the EVRP - limited vehicle capacity and limited vehicle range. 
This reflects the current state of research on the EVRP - many specific constraints and extensions to the EVRP were considered in the literature cited earlier.
However, very few authors focused on solving this fundamental variant, which can be seen as a combination of the Capacitated VRP and a simple variant of the Green VRP (\textcite{ERDOGAN2012100}).
The competition formulation was first introduced in~\textcite{gonccalves2011optimization}, and only a few authors addressed it.
Most of them call the problem Capacitated Green Vehicle Routing Problem (CGVRP) and often consider it an additional constraint on the maximal fleet size. 
Probably the first one is~\textcite{Zhang2018}, who reintroduced the problem to the operations research community and proposed two solution methods: a simple two-phase constructive heuristic and the Ant Colony System (ACS) metaheuristic.
The proposed two-phase heuristic constructs a tour that visits all customers using the Nearest Neighbor heuristic in the first phase and transforms the tour into a valid CGVRP one in the second phase, with a newly proposed repair procedure.
The second proposed method is the ACS adapted for EVRP, a bioinspired algorithm that mimics ant pheromone-based communication (first introduced in~\textcite{Colorni91}).
Both methods are tested on an artificial dataset of randomly generated instances.
The same dataset was then used by~\textcite{Wang2019}, who proposed a Memetic Algorithm and outperformed the method of~\textcite{Zhang2018}.
Then,~\textcite{Normasari2019} proposed a Simulated Annealing metaheuristic for the same problem, with an additional constraint of maximum travel time.
The authors did not compare with other methods and created a custom dataset based on the GVRP dataset used in~\textcite{ERDOGAN2012100}.
Thus, this method is the only neighborhood-oriented metaheuristic deployed on an EVRP variant similar to the competition formulation.
\begin{wrapfigure}{r}{0.5\textwidth}
\centering{\includegraphics[width=0.5\textwidth]{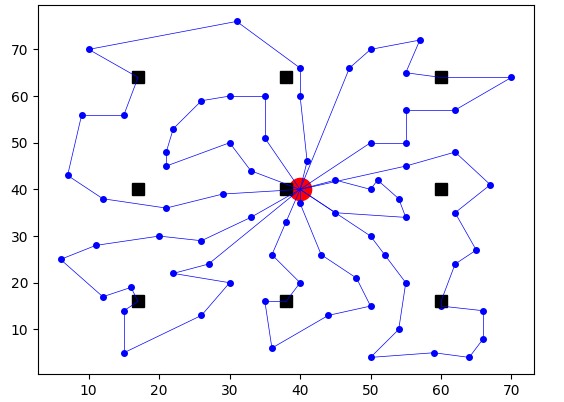}}
   \caption{Solved EVRP instance E-n76-k7}
   \label{fig:solved_problem}
\end{wrapfigure}
Then, there are the five currently unpublished methods submitted to the competition, with results publicly available at~\textcite{competition}: a Sequence-based Hyperheuristic (Kheiri), Simulated Annealing (Mak-Hau and Hill), Max-Min Ant System (Leite, Bernar\-dino, and Goncalves), Genetic Algorithm (Hien, Dao, Thang, and Binh) and Variable Neighborhood Search described in this paper. 
As the problem formulation and stop condition are identical for all of these methods and the dataset is publicly available, the competition results are ideal for comparison with the largest portion of methods proposed for this variant of the EVRP.
Finally,~\textcite{Jia2021} proposed a Bilevel Ant Colony Optimization (BACO) algorithm after the competition and tested it on the competition dataset, using the same problem formulation.
BACO is also included in the comparison presented in this paper.

\section{Problem Formulation}\label{sec:prob_form}

The EVRP is an extension of the classical VRP problem.
The goal of the EVRP is identical to the VRP, finding a set of routes for a fleet of vehicles such that the total distance traveled is minimal.
However, the EVRP introduces two additional constraints.
First, each customer has a specific demand that must be satisfied in a single visit.
Second, electric vehicles (EVs) have a limited battery charge level that must not fall below zero. 
All EVs start at a single central depot and can recharge their batteries at multiple AFSs (alternative fuel stations) and the depot.
 EVs always leave the AFS fully charged and the depot fully charged and loaded.
An example of a solved EVRP instance is shown in Figure~\ref{fig:solved_problem}.
Here, the depot, the AFSs, and the customers are represented by a red circle, black squares, and blue circles, respectively.
The blue line represents the planned EVRP tour.

The mathematical formulation of EVRP as introduced in~\textcite{tech_report} follows.
\begin{equation}\label{eq:obj}
    \min \sum \limits_{i \in V, j \in V, i \neq j} w_{ij}x_{ij},
\end{equation}
s.t.
\begin{equation}\label{eq:cust_connect}
    \sum \limits_{j \in V, i \neq j} x_{ij}=1, \forall i \in I, 
\end{equation}
\begin{equation}\label{eq:afs_connect}
    \sum \limits_{j \in V, i \neq j} x_{ij} \leq 1, \forall i \in F',
\end{equation}
\begin{equation}\label{eq:flow_conservation}
    \sum \limits_{j \in V, i \neq j} x_{ij} - \sum \limits_{j \in V, i \neq j} x_{ji}=0, \forall i \in V, 
\end{equation}
\begin{equation}\label{eq:demand_1}
    u_j \leq u_i - b_i x_{ij} + C(1 - x_{ij}), \forall i \in V, \forall j \in V, i \neq j,
\end{equation}
\begin{equation}\label{eq:demand_2}
    0 \leq u_i \leq C, \forall i \in V,
\end{equation}
\begin{equation}\label{eq:battery_1}
    y_j \leq y_i - hw_{ij}x_{ij} + Q(1 - x_{ij}), \forall i \in I, \forall j \in V, i \neq j,
\end{equation}
\begin{equation}\label{eq:battery_2}
    y_j \leq Q - hw_{ij}x_{ij}, \forall i \in F' \cup D, \forall j \in V, i \neq j,
\end{equation}
\begin{equation}\label{eq:battery_3}
    0 \leq y_i \leq Q, \forall i \in V,
\end{equation}
where $V=\{D \cup I \cup F'\}$ is a set of nodes.
$D$ denotes the central depot, $I$ the set of customers, and $F$ the set of unique AFSs.
To simplify the modeling of the problem, let $F'$ be the set containing $\beta$ copies of each AFS $i \in F$ (i.e., $|F'|=\beta|F|$).
The upper bound $\beta=2|I|$, since each AFS $i \in F$ might need to be visited before and after serving each customer, and each AFS copy must be visited at most once.
Let us define $E=\{(i,j)|i, j \in V, i \neq j\}$ as a set of edges in the fully connected weighted graph $G=(V,E)$.
Then, $x_{ij} \in \{0, 1\}$ is a binary decision variable that corresponds to the usage of the edge from node $i \in V$ to node $j \in V$, and $w_{ij}$ is the weight of this edge.
The competition defines the weight $w_{ij}$ as the Euclidean distance between the nodes $i$ and $j$.
Therefore, the EVRP is both metric and symmetric.
The variables $u_i$ and $y_i$ denote, respectively, the remaining carrying capacity and the remaining battery charge level of an EV on its arrival at node $i \in V$.
Finally, the constant $h$ denotes the consumption rate of the EVs, $C$ denotes their maximum carrying capacity, $Q$ the maximum level of battery charge, and $b_i$ the demand of each customer $i \in I$. 

Then, the individual equations have the following meaning.
Equation~\ref{eq:obj} defines the objective function.
Equation~\ref{eq:cust_connect} enforces visiting each customer exactly once, and Equation~\ref{eq:afs_connect} limits visiting each AFS copy at most once.
Equation~\ref{eq:flow_conservation} ensures the equality between the number of incoming and outgoing arcs at each node.
Equations~\ref{eq:demand_1} and~\ref{eq:demand_2} guarantee fulfilling customer demands and respecting vehicle load capacity.
Equations~\ref{eq:battery_1} and~\ref{eq:battery_2} control battery charge consumption and recharging, while Equation~\ref{eq:battery_3} ensures respecting battery capacity.

For the purposes of formal algorithm description, let us also define an EVRP tour $T$ as a sequence of nodes $T = \{v_0, v_1, ..., v_{n-1}\}$, where $v_i$ is a customer, a depot or an AFS and $n$ is the length of the tour $T$.
Finally, let
\begin{equation}
    w(T) = \sum \limits_{i = 0}^{n - 2} w_{i, i + 1}
\end{equation}
be the weight of the whole tour $T$.

\section{Method}\label{sec:methods}

This chapter is structured as follows.
Section~\ref{sec:vns} introduces the general VNS metaheuristic.
Section~\ref{construction} describes the initial solution construction used within the VNS.
The construction consists of three phases: the Density-Based Clustering Algorithm used for the spatial clustering of customers (Section~\ref{sec:clustering}), the Modified Clarke-Wright Savings Algorithm for creating a load-capacitated tour from the clusters (Section~\ref{sec:clarke-wright}), and the Relaxed ZGA repair procedure for fixing potentially violated battery constraints (Section~\ref{sec:ZGA}).
Then, the perturbation operator used within the VNS is described in Section~\ref{perturbation}.
The rest of the chapter is devoted to the local search part of the VNS.
Section~\ref{sec:local_search} introduces the Randomized Variable Neighborhood Descent heuristic used to control the local search, and Section~\ref{sec:operators} presents the individual operators used to search for different neighborhoods.

\subsection{The approach}\label{sec:vns}

VNS is a metaheuristic method proposed by~\textcite{MLADENOVIC19971097} and is commonly used to approximate the solutions to optimization problems.
It systematically alternates two mechanisms: an exhaustive local search and a perturbation.
The local search reaches a local optimum with respect to all available neighborhoods, whereas the perturbation attempts to escape the valley corresponding to the optimum, thus enabling it to reach a different one.
The process is described in Algorithm~\ref{alg:vns}.

\begin{algorithm}[h]
\SetAlgoLined
\KwIn{EVRP instance}
\KwOut{near-optimal valid EVRP tour $T^{**}$}
 $T^{**} \gets \emptyset $\\
 \While{\bf{not} $\mathtt{stop()}$}{    \label{vns:0}
    $i \gets 0$ \\
    $T^* \gets \mathtt{construction()}$ \\ \label{vns:1}
    \While {$($\bf{not} $\mathtt{stop()}$ \bf{and} $\mathtt{i < ITERS\_MAX}$)}{ \label{vns:1.1}
        $i \gets i+1$ \\
        $T \gets \mathtt{perturbation}(T^*)$ \\ \label{vns:2}
        $T \gets \mathtt{local\_search}(T)$ \\ \label{vns:2.1}
        \If{$w(T) < w(T^*)$  }{ \label{vns:3}
            $T^* \gets T$\\
            $i \gets 0$ \label{vns:3.1}
        }
    }
    \If{$T^{**} = \emptyset$ \bf{or} $(w(T^*) < w(T^{**})$)}{ \label{vns:3.2}
        $T^{**} \gets T^*$\\    \label{vns:4}
    }
 }
 \Return{$T^{**}$}
 \caption{Variable Neighborhood Search (VNS)}\label{alg:vns}
\end{algorithm}

The outer loop (lines~\ref{vns:0}-\ref{vns:4}) is controlled by an arbitrary stop condition $\mathtt{stop()}$ and initializes full restarts of the VNS metaheuristic, while maintaining the overall best solution $T^{**}$.
Each restart starts with the construction of an initial solution $T^*$, using the method described in Section~\ref{construction} (line~\ref{vns:1}).
Then, the following process repeats in the inner loop (lines~\ref{vns:1.1}-\ref{vns:4}).
The current best solution $T^*$ is modified by a randomized perturbation, described in Section~\ref{perturbation}.
The perturbation returns a possibly non-improving current solution $T$ (line~\ref{vns:2}).
The current solution $T$ is then subject to a systematic local search (line~\ref{vns:2.1}), described in Section~\ref{sec:local_search}.
If the cost of the current solution $ w(T)$ after local search is better than $ w(T^*)$, $T$ replaces $T^*$ as the new current best solution and the counter $i$ is reset (lines~\ref{vns:3}-\ref{vns:3.1}).
Similarly, the current best solution $T^*$ is compared to the overall best solution $T^{**}$, and if it is better, $T^*$ replaces $T^{**}$ (lines~\ref{vns:3.2}-\ref{vns:4}).
The inner loop repeats if $\mathtt{stop()}$ is not met and the number of consecutive non-improving iterations is lower than $\mathtt{ITERS\_MAX}$.

The EVRP competition rules define the stop condition $\mathtt{stop()}$ as the maximum number of evaluations of the fitness function $\mathtt{EVALS\_MAX}=25000 \times n$.
Alternatively, maximum running time is used in some experiments.
The maximum number of consecutive non-improving iterations is defined as $\mathtt{ITERS\_MAX}=r \times n$, where $r$ is a restart parameter.

The individual components of the VNS are described in the following sections.
Time complexity is discussed for those components that are used within the inner VNS loop, as these have a dominant effect on the total algorithm time complexity.
The complexity of the construction procedure has a negligible effect on the algorithm performance since it is called only every $\mathtt{ITERS\_MAX}$ iteration.

\subsection{Initial solution construction}\label{construction}

A valid initial solution is constructed before entering the inner VNS loop.
As will be shown in Section~\ref{sec:results}, the quality of the initial construction influences the quality of the final solution.
Based on the preliminary study (\textcite{Kozak2021}), the most promising components were identified and used to create a total number of 16 construction procedures.
All of these were tested within the VNS metaheuristic, and then the best-performing setup was chosen.
It consists of three phases, described briefly in the following sections.
First, all customers are divided into clusters by the Density-Based Clustering Algorithm (Section~\ref{sec:clustering}).
Clustering identifies subsets of customers close to each other, as they are likely to be served by the same EV in an optimal tour.
Second, each of these clusters is transformed into one or more capacitated subtours by the Modified Clarke-Wright Savings Algorithm (Section~\ref{sec:clarke-wright}).
This algorithm initially considers a separate subtour for each customer; therefore, load capacity constraints cannot be violated.
Then, the algorithm joins these subtours according to a simple saving rule.
This rule guarantees that the total distance traveled is maintained or decreased and does not violate the load constraints already fulfilled.
Finally, potential violations of battery constraints are fixed by the Relaxed ZGA repair procedure (Section~\ref{sec:ZGA}).
Thus, a valid EVRP tour is constructed.

\subsubsection{Density-Based Clustering Algorithm (DBCA)}\label{sec:clustering}

The algorithm was first introduced in ~\textcite{Ester96} as Density-Based Spatial Clustering of Applications with Noise (DBSCAN).
The DBCA divides customers from $I$ into disjunctive clusters, defined by two parameters: neighborhood radius $\epsilon$ and density threshold $\delta$.
Let us define the $\epsilon$-neighborhood $N_{\epsilon}(n_i)$ of a node $n_i$ as the set of nodes within $\epsilon$ distance from $n_i$.
Let us also define the condition for the initial cluster size as
\begin{equation} \label{eq:dbca_cond}
|N_{\epsilon}(n_{i})| \geq \delta.    
\end{equation}

\begin{figure}[h]
    \begin{subfigure}[b]{0.5\textwidth}
    \centering
    \includegraphics[width=0.95\textwidth]{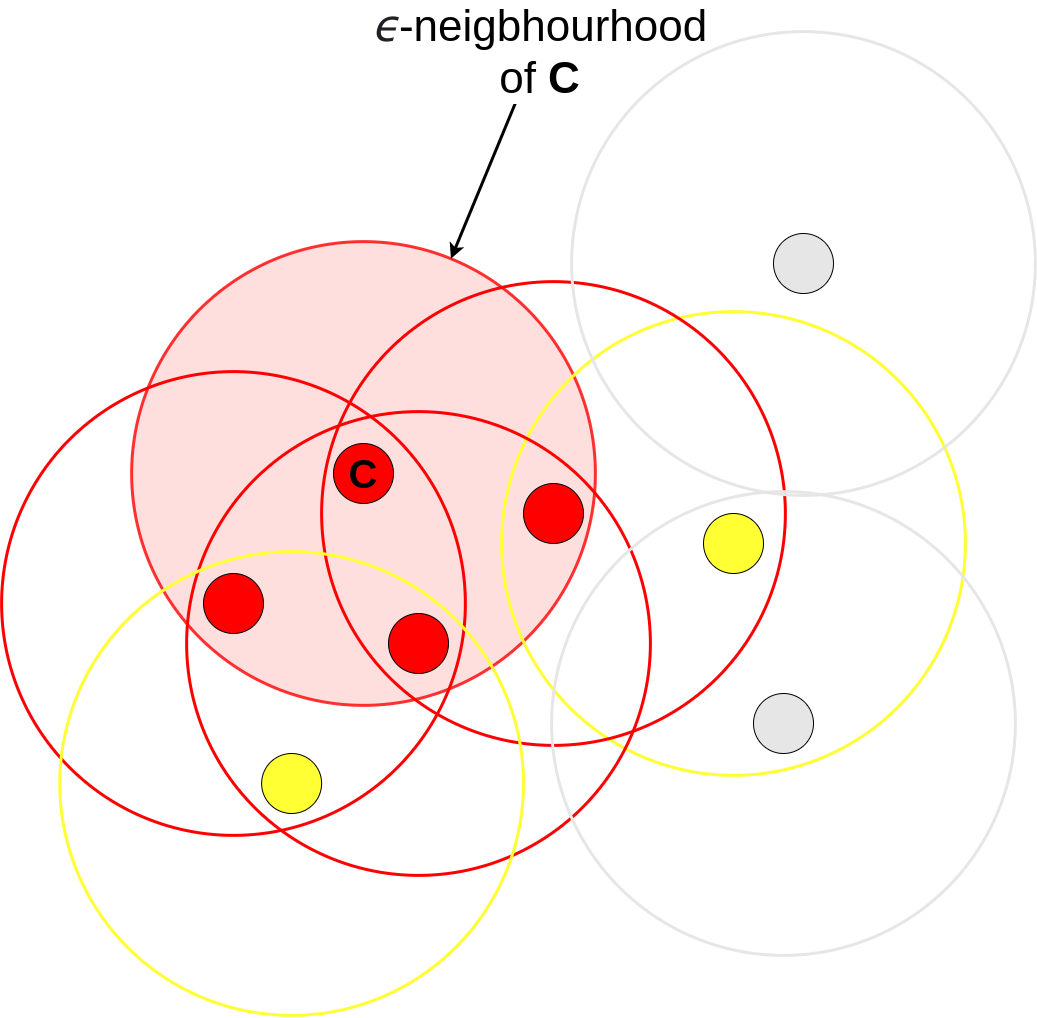}
    \caption{Node classes}
    \label{fig:dbca}
    \end{subfigure}
    \hfill
    \begin{subfigure}[b]{0.5\textwidth}
    \centering
    \includegraphics[width=0.95\textwidth]{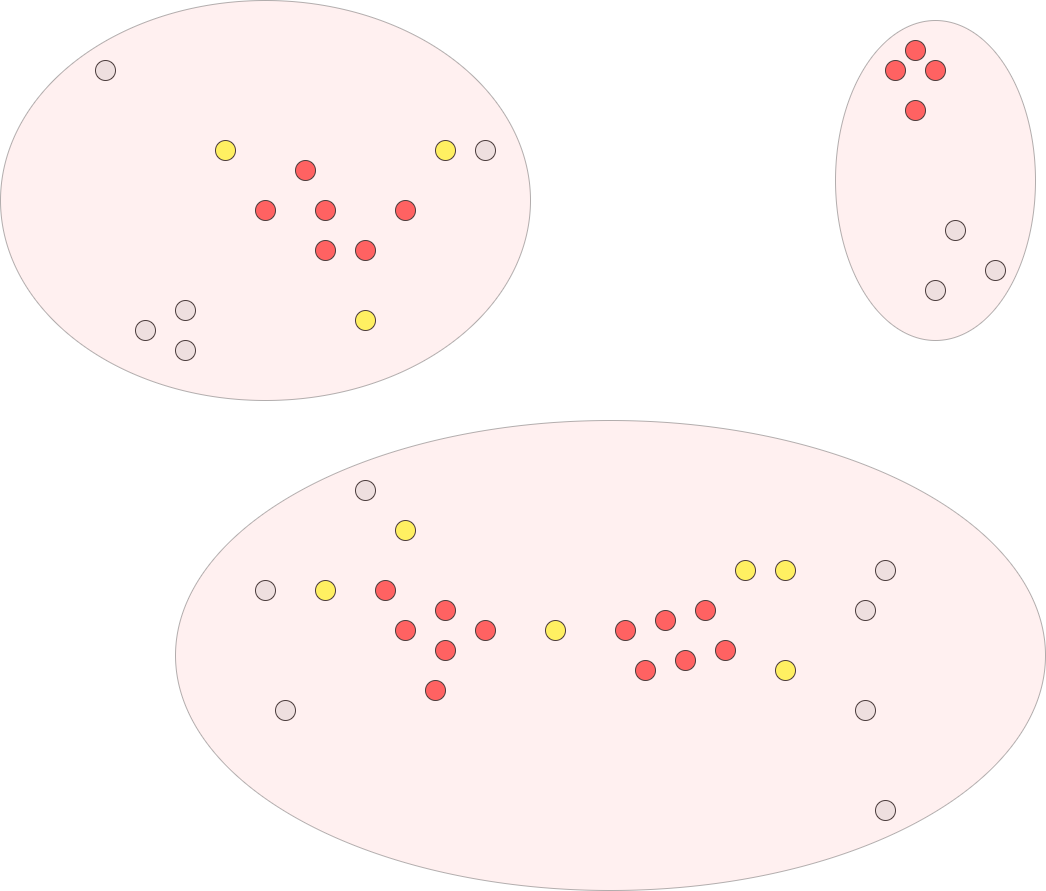}
    \caption{Result of clustering}
    \label{fig:dbca_multi}
    \end{subfigure}
    \caption{Density Based Clustering Algorithm (DBCA)}
\end{figure}

These terms can be used to distinguish three classes of nodes: cluster cores, border nodes, and noise nodes illustrated in Figure~\ref{fig:dbca}.
Cluster cores (red) are nodes $n_i$, for which the condition of Equation~\ref{eq:dbca_cond} holds.
Border nodes (yellow) are nodes within $\epsilon$ neighborhood of a cluster core but are not a cluster core themselves.
Noise nodes (grey) are not in an $\epsilon$-neighborhood of any cluster core.
The $\epsilon$-neighborhood corresponds to the circle around a node.
Red circles are neighborhoods of cluster cores - these neighborhoods contain at least $\delta=4$ nodes, including the cluster core.

The DBCA works as follows.
First, all customers are classified into the three introduced classes.
Then, initial clusters are created as cluster cores, together with nodes within their $\epsilon$-neighborhood.
After that, all non-disjunctive initial clusters are merged.
Finally, the noise nodes are assigned to the clusters closest to them, and thus all nodes are split into disjunctive clusters.
An example of multiple clusters generated by the DBCA is shown in Figure~\ref{fig:dbca_multi}.

The most suitable choice of parameters $\epsilon$ and $\delta$ varies significantly between individual instances.
In order to overcome this issue, these two parameters are determined at runtime, separately, for each problem instance.
Clustering is performed for each combination of the values of $\epsilon$ and $\delta$ given in Table~\ref{tbl:dbc_parameters}, which were empirically preselected.
After that, the best tour obtained after applying the two further described repair procedures is chosen as the initial solution.
The parameter $\epsilon$ is defined as a fraction of $reach$, which is equivalent to the operating range of an EV without recharging.

\begin{table}[ht]
\centering
 \begin{tabular}{cc} 
 \hline
 Parameter & Values \\ \hline
 $\epsilon$ & [$\frac{1}{2}$, $\frac{1}{3}$, $\frac{1}{4}$, $\frac{1}{6}$ 
 , $\frac{1}{8}$, $\frac{1}{10}$, $\frac{1}{15}$, $\frac{1}{20}$] $\times reach$ \\ 
 & \\
 $\delta$ & 2, 3, 4, 5 \\
 \hline
\end{tabular}
\caption{Parameter values for the DBCA}\label{tbl:dbc_parameters}
\end{table}

\subsubsection{Modified Clarke-Wright Savings Algorithm (MCWSA)}\label{sec:clarke-wright}

MCWSA is an evolution of the Clarke-Wright heuristic for the VRP, first introduced in~\textcite{clarke64}.
The modified version presented in~\textcite{ERDOGAN2012100} adapts the original heuristic for the CVRP.
Here, it is applied separately to the clusters created by the DBCA.
These clusters are transformed into valid CVRP tours, which are then simply merged into a single CVRP tour that visits all customers.

In the following description, a tour $T$ is assumed to be stored as a variable-sized array of nodes.
The operation $T.\mathtt{push\_front(i)}$ inserts a node $i$ into the first position of $T$, $T.\mathtt{front()}$ returns the first node of $T$ and $T.\mathtt{pop\_front()}$ returns the first node of $T$, while also removing it from $T$.
The operations $T.\mathtt{push\_back(i)}, T.\mathtt{back()}$ and $T.\mathtt{pop\_back()}$ operate analogously, but with the last node of $T$.
All of these operations have a constant time complexity.

The Algorithm~\ref{alg:MCWSA} describes the processing of a single cluster by the MCWSA.
The input is a subset of customers $I_c$, which corresponds to the processed cluster, the depot $D$, and the maximum vehicle load capacity $C$.
The output is a valid CVRP tour $T_c$.
$T_c$ is initialized by adding the depot $D$ (line~\ref{mcwsa:1}).

The inner loop (lines~\ref{mcwsa:2.5}-\ref{mcwsa:5}) creates exactly one subtour $T_{tmp}$, corresponding to individual round trip from $D$ to $D$.
The outer loop (lines~\ref{mcwsa:1.1}-\ref{mcwsa:7}) restarts the inner loop if more than one round trip is needed to satisfy all customers from $I_c$.
In each restart, the farthest customer from the depot is marked as $next$ and the current vehicle load capacity $C_{tmp}$ is initialized (lines~\ref{mcwsa:1.9}-\ref{mcwsa:2}).
Then, the following process is repeated in the inner loop until the vehicle load capacity $C_{tmp}$ is not sufficient to satisfy the customer $next$.
If the node $closest$ is initialized and equal to the first node of $T_{tmp}$, $next$ is added at the beginning of $T_{tmp}$. 
Otherwise, it is added at the end.
Then $next$ is removed from $I_c$, and $C_{tmp}$ is decreased accordingly (lines~\ref{mcwsa:3}-\ref{mcwsa:4}).
After that, a new $closest$ from the pair $\{T_{tmp}\mathtt{.front()}, T_{tmp}\mathtt{.back()}\}$ and a new $next$ from $I_c$ are determined, so the insertion cost of adding $next$ at the beginning or at the end of $T_{tmp}$ is minimal (line~\ref{mcwsa:5}).
When the inner loop ends, a depot is added at the end of $T_{tmp}$ and $T_{tmp}$ is appended to $T_c$ (lines~\ref{mcwsa:6}-\ref{mcwsa:7}).
The outer loop terminates when all customers in the cluster $I_c$ are visited in $T_c$.

\begin{algorithm}[ht]
\SetAlgoLined
\SetKwInOut{Input}{input}
\newcommand\mycommfont[1]{\ttfamily{#1}}
\SetCommentSty{mycommfont}
\SetKw{Break}{break}
\KwIn{subset of customers $I_c$, depot $D$, vehicle load capacity $C$}
\KwOut{valid CVRP tour $T_c$}
    $T_c\mathtt{.push\_back(}D\mathtt{)}$ \\        \label{mcwsa:1}
    \While{\textup{$|I_c| > 0$}}{   \label{mcwsa:1.1}
        $T_{tmp} \gets \emptyset$ \\
        $closest \gets \emptyset$ \\
        $next \gets \underset{i \in I_c} \arg\max w_{D,i}$ \\   \label{mcwsa:1.9}
        $C_{tmp} \gets C$ \\    \label{mcwsa:2}
        \While{$(C_{tmp} - \mathtt{demand(}next\mathtt{)} \geq 0$ \bf{and} $|I_c| > 0)$}{   \label{mcwsa:2.5}
            \If{$closest = T_{tmp}\mathtt{.front()}$}{          \label{mcwsa:3}
                $T_{tmp}\mathtt{.push\_front(}next\mathtt{)}$ \\
            } 
            \Else{
                $T_{tmp}\mathtt{.push\_back(}next\mathtt{)}$ \\                
            }
            $C_{tmp} \gets C_{tmp} - \mathtt{demand(}next\mathtt{)}$ \\
            $I_{c}\mathtt{.remove(}next\mathtt{)}$ \\           \label{mcwsa:4}
            $next, closest \gets \underset{i \in I_c, j \in \{T_{tmp}\mathtt{.front()}, T_{tmp}\mathtt{.back()}\}} {\arg \min (w_{i,j} - w_{D, j} - w_{D, i})}$ \\ \label{mcwsa:5}
        }
        $T_{tmp}\mathtt{.push\_back(}D\mathtt{)}$ \\        \label{mcwsa:6}
        $T_c\mathtt{.append(}T_{tmp}\mathtt{)}$         \label{mcwsa:7}
    }
    \Return{$T_c$}
\caption{Modified Clarke-Wright Savings Algorithm (MCWSA)}\label{alg:MCWSA}
\end{algorithm}

Let us consider a tour $T_{tmp}$, a tour $T_{next}$, which visits only the node $next$ on a round trip from the depot and back, and a tour $T_{tmp} \cup next$, which is a tour created by appending $next$ to $T_{tmp}$ within the MCWSA.
Note that the MCWSA guarantees $w(T_{tmp} \cup next) \leq w(T_{tmp}) + w(T_{next})$,
if the triangle inequality holds.
Together with its low complexity, this property is the main benefit of the MCWSA.

\subsubsection{Relaxed ZGA repair procedure}\label{sec:ZGA}

The Relaxed ZGA repair procedure is the last step of the construction process, as it transforms a CVRP tour produced by the MCWSA into a valid EVRP tour.
In addition, it is a universal repair procedure capable of fixing any violated constraints.
This property is utilized in the perturbation (Section~\ref{perturbation}) and the local search operators reallocating AFSs (Section~\ref{sec:operators}).
The repair procedure is a one-pass algorithm.
It sequentially copies the customer nodes from the input tour to the output tour while performing a sequence of constraint checks and inserting additional visits to the AFSs or the depot.

 Relaxed ZGA is based on a method introduced in~\textcite{Zhang2018}, and ZGA is an acronym of the authors' names.
As was shown in~\textcite{Kozak2021}, the original ZGA method can fail to produce a valid EVRP tour, even if one exists.
Therefore, the method was modified and named Relaxed ZGA, which is the variant presented in Algorithm~\ref{alg:zga}.
It is assumed that all customers are within half of the battery range of some AFS and that any AFS is reachable from any other AFS.

The input of the method is a tour $T$, which must start and end at the depot and can visit any sequence of nodes.
The output is a valid EVRP tour $T_{EVRP}$.
In the algorithm, $current$ refers to the last node added to $T_{EVRP}$ and $next$ to the first node in $T$ not yet added.
The $T_{EVRP}$ is initialized by adding the depot $D$, which is also set as $current$. (lines~\ref{zga:1}-\ref{zga:2}).
Then, the nodes from $T$ are sequentially added to $T_{EVRP}$, until there is no node left in $T$ (lines~\ref{zga:3}-\ref{zga:5}).
If $next$ cannot be satisfied due to insufficient remaining load capacity, $next$ is set to the depot instead (lines~\ref{zga:3.5},~\ref{zga:4}-\ref{zga:5}).
Otherwise, the remaining vehicle range after visiting $next$ is checked.
If the vehicle can still reach an AFS after serving $next$, it can be safely added to $T_{EVRP}$.
Otherwise, the vehicle is sent to the AFS closest to $next$ that is reachable from $current$ (lines~\ref{zga:3.6}-\ref{zga:7}).

\begin{algorithm}[ht]
\SetAlgoLined
\SetKwInOut{Input}{input}
\newcommand\mycommfont[1]{\ttfamily{#1}}
\SetCommentSty{mycommfont}
\SetKw{Break}{break}
\KwIn{any tour $T$, depot $D$}
\KwOut{valid EVRP tour $T_{EVRP}$}
$T_{EVRP}\mathtt{.push\_back(}T\mathtt{.pop\_front())}$   \\  \label{zga:1}
$current \gets T_{EVRP}\mathtt{.back()}$                    \\ \label{zga:2}
$next \gets T\mathtt{.front()}$                       \\  
\While{\textup{$T$ not empty}}{                         \label{zga:3}
    \If{$\mathtt{remaining\_load(}T_{EVRP}\mathtt{)} \geq \mathtt{demand(}next\mathtt{)}$}{ \label{zga:3.5}
        \If{\textup{AFS can be reached via $next$}}{    \label{zga:3.6}
            $T_{EVRP}\mathtt{.push\_back(}next\mathtt{)}$       \\  \label{zga:8}
            $current \gets next$    \\
            $T\mathtt{.pop\_front()}$                 \\
            $next \gets T\mathtt{.front()}$               \label{zga:9}
        }
        \Else{
            $next \gets $ AFS closest to $next$ reachable from $current$\\ \label{zga:6}
            $T\mathtt{.push\_front(}next\mathtt{)}$                \\ \label{zga:7}
        }
    }
    \Else{
        $next \gets D$ \label{zga:4} \\
        $T\mathtt{.push\_front(}next\mathtt{)}$ \label{zga:5} 
    }
}
\Return{$T_{EVRP}$}
\caption{Relaxed ZGA repair procedure}\label{alg:zga}
\end{algorithm}

As for time complexity, the Relaxed ZGA procedure iterates once over the input tour, which visits some customers from $I$, some AFSs from $F$, and the depot $D$.
Given that each customer can require at most one visit to the depot and that any two customers are at most two AFSs away (the second Relaxed ZGA assumption), this outer loop iterates over $k \times n$ nodes, where $n=|V|$ is the size of the EVRP instance, and $k$ is a constant.
The test of whether the vehicle will get stuck due to a low battery level is performed for each added node, and the AFS closest to $next$ still reachable from $current$ is determined.
Both of these operations can be carried out within one pass over all AFSs in $F$.
Therefore, the whole complexity of Relaxed ZGA can be estimated as $\mathcal{O}(n \times |F|)$.

\subsection{Local Search}\label{sec:local_search}

Efficient local search is an essential part of the VNS metaheuristic.
The local search is realized by iteratively performing the most improving modification of the current solution in some local neighborhoods.
When no improving move is available, a local optimum is reached w.r.t. the neighborhood.
It is advantageous to search through more than one neighborhood, as the local optimum w.r.t. one neighborhood may be further improved in another one.
In the VNS, individual neighborhoods correspond to local search operators applicable to the current solution.
Some of the operators used (2-opt and 2-string) are commonly used permutation-based operators that were already implemented in our previous GRASP method (\textcite{Woller2021}).
The remaining operators (AFS-realloc-1 and AFS-realloc-more) are problem-specific and are newly added to the VNS.

The local search is controlled by a simple heuristic called Randomized Variable Neighborhood Descent (RVND, \textcite{Duarte2018}).
This heuristic was already used in the previous GRASP method and is commonly used in various metaheuristics.
The heuristic searches for the local optimum sequentially in different neighborhoods, in random order.
Once an improving solution is found, the search is restarted, and the order of the neighborhoods is randomly reshuffled.
If no improvement is achieved in any of the neighborhoods, the RVND is terminated.
A more detailed description can be found in~\textcite{Woller2021}.
The rest of this section provides a description of the individual local search operators.

\subsubsection{Local Search Operators}\label{sec:operators}

This section describes individual local search operators.
Two problem-specific operators are proposed: AFS-realloc-1 (based on the Recharge Relocation operator from~\textcite{Felipe2014}) and AFS-realloc-more.
In addition to these, two more operators commonly used in permutation-based optimization problems are adapted for the EVRP.
The first is the well-established 2-opt (\textcite{Englert2014}).
The second one, called 2-string (~\textcite{Mikula2022cejor}), is a generalization of numerous other commonly used operators, such as 2-exchange, swap, and relocate.
Permutation-based operators were already used in our previous GRASP method.
For this reason, they are only briefly introduced here, and a detailed description can be found in~\textcite{Woller2021}.

\begin{figure}[ht]
    \begin{subfigure}[b]{0.5\textwidth}
    \centering
    \includegraphics[width=0.95\textwidth]{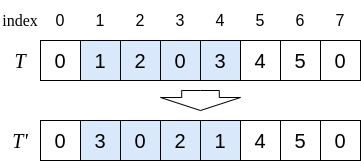}
    \caption{$T'$=2-opt($i=1, j=4, T$)}
    \label{fig:two_opt}
    \end{subfigure}
    \hfill
    \begin{subfigure}[b]{0.5\textwidth}
    \centering
    \includegraphics[width=0.95\textwidth]{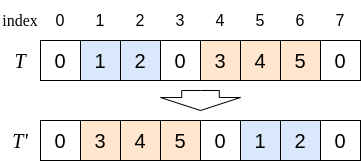}
    \caption{$T'$=2-string$(i=1, j=4, X=2, Y=3, T)$}
    \label{fig:two_string}
    \end{subfigure}
    \caption{Permutation-based operators}
    \vspace{-1.5em}
\end{figure}

\vspace{-0.5em}
\paragraph{2-opt} 2-opt takes three parameters as input: two indices $i, j$, and a tour $T$ and returns a modified tour $T'$, where the sequence of nodes from the index $i$ to the index $j$ is reversed.
It must hold that $i < j$, $i \geq 0$ and $j < n$.
The 2-opt operator is illustrated in Figure~\ref{fig:two_opt}.


\paragraph{2-string and its variants} 2-string takes five parameters as input: a tour $T$, a pair of indices $i, j$ valid w.r.t. to $T$, and a pair of non-negative integers $X, Y$.
It returns a modified tour $T'$, where the sequence of $X$ nodes starting with the $i$-th node in $T$ is swapped with the sequence of $Y$ nodes starting with the $j$-th node.
It must hold, that $i \geq 0$, $j \geq i + X$ and $j + Y < n-1$.
The 2-string operator is illustrated in Figure~\ref{fig:two_string}.
The following seven operators were created by fixing the values of $X$ and $Y$:

\begin{multicols}{2}
\begin{itemize}
    \item 1-point: $X=0, Y=1$
    \item 2-point: $X=1, Y=1$
    \item 3-point: $X=1, Y=2$
    \item or-opt2: $X=0, Y=2$
    \item or-opt3: $X=0, Y=3$
    \item or-opt4: $X=0, Y=4$
    \item or-opt5: $X=0, Y=5$
\end{itemize}
\end{multicols}

When performing the local search, the complementary variants of these operators (e.g., 1-point with $X=1, Y=0$) are also considered.

\paragraph{AFS-realloc-1}

\begin{wrapfigure}{r}{0.4\textwidth}
\centering{\includegraphics[width=0.4\textwidth]{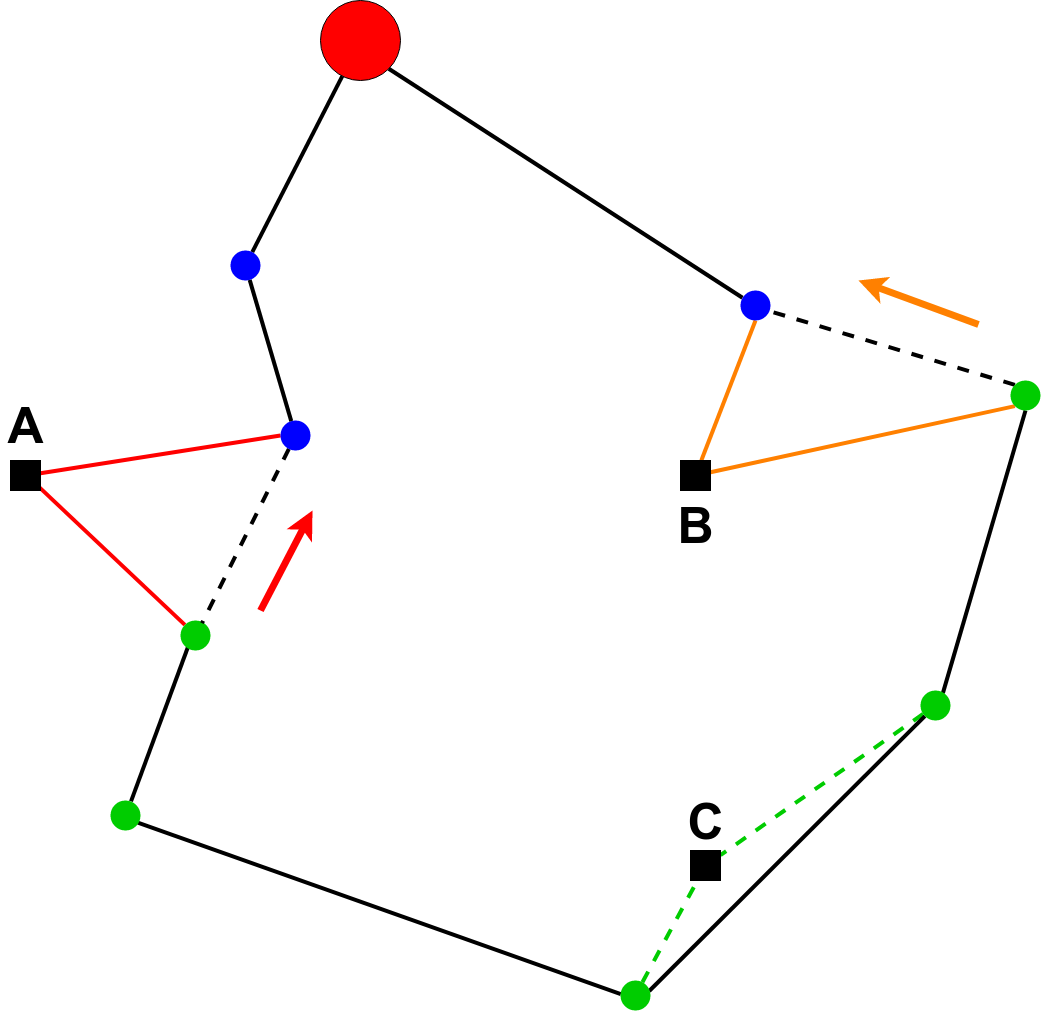}}
   \caption{AFS-realloc-1}\label{fig:reallocation}
\end{wrapfigure}

An EVRP tour $T$ can be divided into smaller subtours, separated by visits to the depot.
The AFS-realloc-1 operator aims to optimally allocate an AFS within a single subtour $t$, while keeping the order of customers untouched.
The optimal reallocation is limited to subtours that visit exactly one AFS due to its acceptable complexity of $\mathcal{O}(n\times|F|)$, where $F$ is the set of all AFSs.
The operator is described in Algorithm~\ref{alg:realloc_1} and illustrated in Figure~\ref{fig:reallocation}.
The following procedure is performed for all subtours $t$ with exactly one AFS.
First, a subtour $t'$ is created by removing the AFS from $t$ (line~\ref{alg:realloc_1.1}).
Then, the last reachable AFS A in $t'$ is determined (line~\ref{alg:realloc_1.2}) by applying the Relaxed ZGA repair procedure described in Section~\ref{sec:ZGA}.
 AFS A is placed so that the maximum number of customers is visited before recharging and the vehicle does not run out of energy before A.
Note that AFS A corresponds to the first usable AFS in a reversed subtour $t'$, so that the vehicle does not run out of energy after visiting it and before reaching the depot.
Similarly, the last AFS B that can be reached is determined for the reversed subtour $t'$ (line~\ref{alg:realloc_1.3}).
The sequence of customers between A and B corresponds to a set of valid insertion edges $E$. The endpoints of these edges are drawn in green in Figure~\ref{fig:reallocation}.
An AFS C can be inserted between any of these customers.
The resulting subtour $t''$ will be valid since the original subtour $t$ was valid and contained only a single AFS.
Note that the direction of the subtour does not matter, as the EVRP is symmetric.
The optimal AFS C to insert is determined by trying all possible AFSs from the set of all AFSs $F$ in combination with all insertion edges from $E$ (lines~\ref{alg:realloc_1.4}-\ref{alg:realloc_1.5}).

An important benefit of the AFS-realloc-1 is that it also considers the AFSs currently unused in the tour $T$.
These are not reachable by the previously described operators 2-string and 2-opt.
Similarly to these, AFS-realloc-1 can be sped up by using an efficient constant-time cost update function $\delta_{AFS-realloc-1}$, instead of determining the total subtour weight $w(t'')$, which is a $\mathcal{O}(n)$ operation.
Let us define the insertion cost of an AFS placed at an index $k$ in a subtour $t$ as
$ I\_cost_{k} = w_{i, k} + w_{j,k} - w_{i,j}.$
Then, the cost update can be evaluated as $ \delta_{AFS-realloc-1} = I\_cost_{original} - I\_cost_{new}, $ where $original$ is an index of the AFS in the subtour $t$ and $new$ of the AFS C inserted to $t''$.

Regarding complexity, AFS-realloc-1 applies the Relaxed ZGA procedure twice to each subtour $t \in T$ with a single AFS.
Therefore, the procedure is performed at most twice for the whole tour $T$.
Similarly, the reallocation loop (lines~\ref{alg:realloc_1.4} to~\ref{alg:realloc_1.5}) is run for all AFSs from $F$ in combination with all insertion edges $e$.
In the worst case, all edges in $T$ are potential insertion edges; thus, the complexity of the reallocation loop is $\mathcal{O}(n \times |F|)$ - the same as for the Relaxed ZGA.
Therefore, the whole complexity of AFS-realloc-1 is also $\mathcal{O}(n \times |F|)$.

\begin{algorithm}[h]
\SetAlgoLined
\SetKwInOut{Input}{input}
\KwIn{valid EVRP tour $T$, set of all AFSs $F$}
\KwOut{potentially improved valid EVRP tour $T$}
\For{\textup{\textbf{each} subtour $t \in T$ with exactly one AFS}}{
$t' \gets \mathtt{removeAllAFSs}(t)$ \\                     \label{alg:realloc_1.1}
$A \gets $ last reachable AFS in $t'$ \\                     \label{alg:realloc_1.2}
$B \gets $ last reachable AFS in $\mathtt{reverse}(t')$ \\   \label{alg:realloc_1.3}
$E \gets $ valid insertion edges between $A$ and $B$ \\
    \For{\textup{\textbf{each}} $e \in E$}{                 \label{alg:realloc_1.4}
        \For{\textup{\textbf{each} AFS} $C \in F$}{
            $t'' \gets t' + C$ at $e$ \\
            \If{$w(t'') < w(t)$}{
                $t \gets t''$                               \label{alg:realloc_1.5}
            }
        }    
    }
}
\Return{$T$}
\caption{AFS-realloc-1}\label{alg:realloc_1}
\end{algorithm}

\paragraph{AFS-realloc-more}
The AFS-realloc-1 is limited to subtours with a single AFS.
Optimally placing more than one AFS would be significantly more complicated, and the complexity would be higher.
For these reasons, the AFS-realloc-more aiming at such subtours is designed only to replace the AFSs used in a subtour and potentially reduce their number.
However, their optimal placement is left to permutation-based operators.
The operator is described in Algorithm~\ref{alg:realloc_more} and works as follows.
For each subtour $t \in T$ with more than one AFS, the following procedure is performed.
First, a subtour $t'$ is created from $t$ by removing all AFSs (line~\ref{alg:realloc_more.1}).
The AFSs are then reinserted into $t'$ by the Relaxed ZGA repair procedure (line~\ref{alg:realloc_more.2}).
The same is done for the reversed subtour $t'$ (line~\ref{alg:realloc_more.3}) and the subtour with the lowest weight is kept in $T$ (lines~\ref{alg:realloc_more.4}-\ref{alg:realloc_more.5}).

\begin{algorithm}[ht]
\SetAlgoLined
\SetKwInOut{Input}{input}
\KwIn{valid EVRP tour $T$}
\KwOut{potentially improved valid EVRP tour $T$}
\For{\textup{\textbf{each} subtour $t \in T$ with more than one AFS}}{
$t' \gets \mathtt{removeAllAFSs}(t)$ \\                     \label{alg:realloc_more.1} 
$t'' \gets \mathtt{Relaxed\_ZGA(t')}$ \\                       \label{alg:realloc_more.2}
$t''' \gets \mathtt{Relaxed\_ZGA(\mathtt{reverse}(t'))}$ \\    \label{alg:realloc_more.3}
\uIf{$w(t'') < w(t''')$ \bf{and} $w(t'') < w(t)$}{             \label{alg:realloc_more.4}
$t \gets t''$
} \uElseIf{$w(t''') < w(t'')$ \bf{and} $w(t''') < w(t)$}{
$t \gets t'''$
}
}
\Return{$T$}                                                \label{alg:realloc_more.5}
\caption{AFS-realloc-more}\label{alg:realloc_more}
\end{algorithm}

The operator complexity is again given by using the Relaxed ZGA repair procedure and is $\mathcal{O}(n \times |F|)$.
No cost update is derived for the AFS-realloc-more operator, as the fitness evaluation is done only twice per subtour, and the dominant operation w.r.t. time demand is the repair procedure.

\subsection{Perturbation}\label{perturbation}

The perturbation is a randomized operation applied to the current best solution $T^*$ within the inner VNS loop (line~\ref{vns:2} of Algorithm~\ref{alg:vns}).
Typically, $T^*$ is a local optimum obtained in a previous iteration of the VNS, and the local search can make no further improvement.
The perturbation is intended to move the search process out of the reach of local search operators while keeping most of the properties of $T^*$.
This follows the assumption of the VNS that local minima with respect to different neighborhoods are relatively similar.
For this purpose, a generalized form of the Double-Bridge perturbation, first introduced in~\textcite{Martin1997} for the Travelling Salesman Problem, is used.
The Generalized Double-Bridge divides $T^*$ into $p+1$ subtours, according to $p$ randomly selected indices.
These subtours are randomly shuffled, inverted, and reconnected, producing a possibly invalid tour.
This tour is then repaired (if necessary) by the Relaxed ZGA repair procedure described in Section~\ref{sec:ZGA} and passed to the local search as a valid tour $T$.

\section{Results and Discussion}\label{sec:results}

In our previous work, we presented a GRASP metaheuristic for the EVRP (\textcite{Woller2021}) and performed a comparison of various initial construction procedures for the CGVRP and related problems (\textcite{Kozak2021}).
For all experiments presented in this paper, the competition dataset is used.
The dataset consists of two sets of problems denoted $E$ and $X$.
The set $E$ is based on the CVRP benchmark set from~\textcite{Christofides69} and contains small instances with at most 100 customers.
The set $X$ is based on instances from~\textcite{Uchoa2016} and contains medium to large instances with 143 to 1000 customers.
The entire competition dataset is publicly available at~\textcite{competition}.
The evolutionary improvement obtained by extending the GRASP is documented in Section~\ref{sec:VNS_evolution}.
The results of the competition of the top three methods are presented in Section~\ref{sec:comp_results}. 
Although these methods have not yet been published, this comparison is valuable, as the competition organizers generated the results with the same stop condition (fixed number of fitness evaluations) and on the same machine.
Finally, Section~\ref{sec:BACO} provides a comparison with the BACO algorithm proposed by~\textcite{Jia2021} after the competition.
\textcite{Jia2021} used a different stop condition, given by a predefined time budget.
The computational budget was significantly larger than in the competition, so the results are not directly comparable.
Therefore, new results for the VNS metaheuristic are generated and presented separately, with the stop condition adequately adjusted. 

The VNS metaheuristic was not directly compared with other methods proposed before the competition, as we did not obtain any of the datasets used in those papers.
Moreover, these methods address a slightly different variant of the EVRP. 
\textcite{Zhang2018} and~\textcite{Wang2019} consider the same constraints as the competition, plus an additional constraint on the maximal fleet size. 
\textcite{Normasari2019} then limits also the maximum travel time.

The tests were carried out on a Linux computer with an Intel Core i7-7700 3.60 GHz CPU.
Both the GRASP and the VNS metaheuristic are implemented in C++.
The implementation of the VNS metaheuristic, as submitted to the competition, is publicly available at~\textcite{vns-evrp-2020_2022}.
All results presented in Section~\ref{sec:VNS_evolution} and Section~\ref{sec:comp_results} are obtained with the same stop condition defined by the competition rules.
The stop condition is defined as the total number of fitness evaluations, and the value is set to $25000n$, where $n=|V|$ is the size of the instance.
Following the competition rules, each instance is solved $\mathtt{RUNS}=20$ times, with seeds $s \in \{1, 2, ..., \mathtt{RUNS}\}$.
Then, the mean score is determined across all runs for all 17 instances from sets $E$ and $X$.
The goal is to obtain the best mean score (BMS) for the highest number of instances.

The metrics shown for each setup in Section~\ref{sec:VNS_evolution}, which describes the evolution of the VNS method, are as follows:
\begin{itemize}
    \item BMS count: number of instances for which the setup obtained the BMS in the current experiment, which is also the primary selection criterion
    \item $\overline{mean\_scores}$: mean scores of the given setup, averaged over all instances from sets $E$ and $X$
    \item BS count: number of instances for which the setup obtained the best score (BS) in the current experiment
    \item $\overline{best\_scores}$: best scores of the given setup, averaged over all instances from sets $E$ and $X$
\end{itemize}
All scores are expressed relative to the GRASP method's best score for the particular instance.
The value of $\overline{mean\_scores}$ is calculated as:

\begin{equation*}
\overline{mean\_scores} = \frac{\sum \limits_{i=1}^{|E \cup X|} \frac{\sum \limits_{s=1}^{\mathtt{RUNS}} \frac{score_{i}^{s}}{ref\_score_{i}}}{\mathtt{RUNS}}}{|E \cup X|}, 
\end{equation*}
where $i$ is the index of an instance, $s$ is a seed, $score_{i}^{s}$ is the score obtained by the current setup in instance $i$ with seed $s$ and $ref\_score_{i}$ is the best score obtained by the GRASP reference method in instance $i$.
Similarly, $\overline{best\_scores}$ is calculated as:
\begin{equation*}
\overline{best\_scores} = \frac{\sum \limits_{i=1}^{|E \cup X|} \frac{best\_score_{i}}{ref\_score_{i}}}{|E \cup X|}, 
\end{equation*}
where $best\_score_{i}$ is the best score obtained on instance $i$ by the current setup across all seeds.

In Section~\ref{sec:BACO}, a predefined time budget $t_{max}$ (in hours) is used as a stop condition:
\begin{flalign}\label{eq:t_max_stop}
  t_{max} = \frac{|I| + |F|}{\mathtt{C}}\nu~[h], 
\end{flalign}
where $|I|$ is the number of customers, $|F|$ is the number of AFSs, $\nu$ equals 1, 2, and 3 for instances E22-E101, X143-X916 and X1001, respectively, and $\mathtt{C} = 100$ is a scaling constant.
It is adopted from~\textcite{Jia2021}, who proposed the Bilevel Ant Colony Optimization (BACO) algorithm for EVRP after the competition.
Direct comparison with BACO on the same machine is not possible because the implementation is not available.
For this reason, the following adjustment of the stop condition is made.
\textcite{Jia2021} generated their results on a Linux computer with an Intel Core i7-6700 3.40 GHz CPU, while the VNS results were generated on a Linux computer with Intel Core i7-7700 3.60 GHz CPU, which is slightly more powerful.
To compensate for this, the VNS was given less time, proportionately to the difference in the performance of both CPUs.
For this purpose, the CPU Single Thread Rating from~\textcite{cpu_benchmark} is used.
The first CPU has a rating $\mathtt{R_1} = 2302$ and the second $\mathtt{R_2} = 2474$. 
Let us denote the time budget used in~\textcite{Jia2021} by $t_{max}^{BACO} = t_{max}$.
Then, the time budget $t_{max}^{VNS}$ given to the VNS is calculated as
\begin{equation}\label{eq:vns_stop}
    t_{max}^{VNS} = \frac{\mathtt{R_1}}{\mathtt{R_2}}t_{max}^{BACO}.
\end{equation}

\subsection{VNS evolution}\label{sec:VNS_evolution}
The reference GRASP metaheuristic consists of an initial solution construction and a local search.
The construction used in the GRASP is a two-phase procedure, which first creates a tour using the Nearest Neighbor (NN) heuristic and then transforms it into a valid EVRP tour with a custom repair procedure called Separate Sequential Fixing (SSF).
The local search is controlled by the RVND heuristic and utilizes only the permutation-based operators: three 2-string variants (1-point, 2-point, 3-point) and 2-opt.

\begin{table}[h]
\begin{tabularx}{\textwidth}{l|l}
\hline
Parameter & Description \\
\hline
ssf         & Separate Sequential Fixing (SSF) repair procedure (\textcite{Woller2021})\\
zga         & Relaxed ZGA repair procedure \\
c           & Index of construction used \\
ls          & Index of problem-specific local search operators subset \\
$p$           & Perturbation parameter \\
$r$           & Restart parameter (VNS is restarted after $r\times n$ nonimproving iterations) \\
\hline
\end{tabularx}
\caption{Parameters description}\label{tab:parameters}
\vspace{-1.5em}
\end{table}

The improvement in performance achieved by each of the extensions added to the GRASP is documented in the following sections.
If some alternative components were tried, their performance is shown as well.
Some settings were adopted from the GRASP without repeated tuning: the subset of permutation-based operators, using Randomized VND in the local search and performing the local search according to the best improvement strategy.
Table~\ref{tab:parameters} explains the parameters and acronyms of the components used in the presentation of the following experiments.
Table~\ref{tab:constructions} lists all the constructions tested.
A detailed description of the constructions tested that were not used in the final VNS can be found in~\textcite{Kozak2021}.

\begin{table}[h]
\begin{tabular}{l|p{0.413\columnwidth}ll}
\hline
Index   & Initialization            & Repair procedure & Seed \\
\hline
c:0     & One Route Each (ORE)      & - & -       \\
c:1-2   & Nearest Neighbour (NN)    & SSF & 1-random, 2-fixed \\
c:3-4   & Random                    & SSF & 3-random, 4-fixed \\
c:5-6   & NN                        & Relaxed ZGA & 5-random, 6-fixed \\
c:7-8   & Random                    & Relaxed ZGA & 7-random, 8-fixed \\
c:9     & Modified Clarke-Wright Savings Algorithm (MCWSA) & SSF & - \\
c:10    & MCWSA & Relaxed ZGA & - \\
c:11    & Density Based Clustering Algorithm (DBCA) + cluster-wise NN & SSF & random \\
c:12    & DBCA + cluster-wise NN & Relaxed ZGA & random \\
c:13    & DBCA + cluster-wise MCSWA & SSF & - \\
c:14    & DBCA + cluster-wise MCSWA & Relaxed ZGA & - \\
c:15    & Minimum Spanning Tree (MST) & SSF & - \\
c:16        & MST & Relaxed ZGA & - \\
\hline
\end{tabular}
\caption{Constructions description}\label{tab:constructions}
\end{table}

\subsubsection{Switching from GRASP to VNS}\label{sec:grasp_to_vns}

Table~\ref{tab:grasp2vns} documents the improvement achieved by replacing the metaheuristic, while the other components remain the same. 
The VNS has two more parameters, whose default values were set as $p=3$, $r=1$ without prior tuning, as they will be tuned in the last step.
The VNS clearly outperforms the GRASP, as it yields a better mean score in 14 out of 17 instances.
Moreover, it is better by 0.9\% in both $\overline{mean\_scores}$ and $\overline{best\_scores}$.

\begin{table}[h]
\begin{tabularx}{\textwidth}{X|rrrr}
\hline
\\[-1em]
Method & BMS count & $\overline{mean\_scores}$ & BS count & $\overline{best\_scores}$ \\
\hline
VNS\_ssf\_c:1\_p:3\_r:1 & 14 & 0.999 & 16 & 0.991 \\
GRASP\_ssf\_c:1 & 3 & 1.008 & 3 & 1.000 \\
\hline
\end{tabularx}
\caption{VNS evolution: Switch from GRASP to VNS}\label{tab:grasp2vns}
\end{table}

\begin{table}[h]
\begin{tabularx}{\textwidth}{X|rrrr}
\hline
\\[-1em]
Method & BMS count & $\overline{mean\_scores}$ & BS count & $\overline{best\_scores}$ \\ 
\hline
VNS\_ssf\_c:14\_p:3\_r:1 & 4 & 1.002 & 6 & 0.993 \\
VNS\_ssf\_c:2\_p:3\_r:1 & 4 & 1.001 & 2 & 0.995 \\
VNS\_ssf\_c:13\_p:3\_r:1 & 3 & 1.001 & 2 & 0.995 \\
VNS\_ssf\_c:1\_p:3\_r:1 & 2 & 0.999 & 4 & 0.991 \\
VNS\_ssf\_c:0\_p:3\_r:1 & 2 & 1.010 & 4 & 1.000 \\
VNS\_ssf\_c:16\_p:3\_r:1 & 1 & 1.004 & 0 & 0.996 \\
VNS\_ssf\_c:6\_p:3\_r:1 & 1 & 1.004 & 1 & 0.997 \\
VNS\_ssf\_c:9\_p:3\_r:1 & 1 & 1.005 & 0 & 0.998 \\
VNS\_ssf\_c:10\_p:3\_r:1 & 0 & 1.003 & 0 & 0.996 \\
VNS\_ssf\_c:11\_p:3\_r:1 & 0 & 1.002 & 3 & 0.992 \\
VNS\_ssf\_c:12\_p:3\_r:1 & 0 & 1.002 & 4 & 0.994 \\
VNS\_ssf\_c:15\_p:3\_r:1 & 0 & 1.005 & 2 & 0.997 \\
VNS\_ssf\_c:3\_p:3\_r:1 & 0 & 1.009 & 1 & 0.999 \\
VNS\_ssf\_c:4\_p:3\_r:1 & 0 & 1.011 & 0 & 1.003 \\
VNS\_ssf\_c:5\_p:3\_r:1 & 0 & 1.003 & 2 & 0.994 \\
VNS\_ssf\_c:7\_p:3\_r:1 & 0 & 1.010 & 0 & 1.000 \\
VNS\_ssf\_c:8\_p:3\_r:1 & 0 & 1.011 & 0 & 1.003 \\
\hline
\end{tabularx}
\caption{VNS evolution: Construction selection}\label{tab:cons_select}
\vspace{-1.5em}
\end{table}

\subsubsection{Construction selection} \label{sec:cons_selection}

Table~\ref{tab:cons_select} shows the performance of the VNS in combination with 17 different variants of a construction procedure (although some variants differ only by fixing the seed, thus producing the same tour each time).
It can be seen that the previous best setup with construction c:1 was outperformed in terms of BMS count only by its deterministic variant and by the two most complex constructions, c:13 and c:14.
Interestingly, the trivial c:0 (construction that makes a separate tour for each customer) does not perform much worse than all other constructions.
Overall, the choice of construction has only a moderate influence on VNS performance, although~\textcite{Kozak2021} documented a great variety in the quality of the solution obtained by different constructions.
A possible explanation is that the stop condition is set rather generously, and the VNS has enough time to escape even low-quality initial solutions. 

\subsubsection{Replacing repair procedure}\label{sec:repl_repair}

In the previous step, the construction c: 14 was selected.
The c:14 is identical to c:13, but it uses the Relaxed ZGA procedure instead of the SSF currently used in the VNS perturbation.
Due to this observation, it was tested to replace the SSF repair procedure also in the perturbation.

Table~\ref{tab:repl_repair} shows that Relaxed ZGA is a better choice for a repair procedure altogether.
Although the mean improvement in both qualitative metrics is less than 0.5\%, using Relaxed ZGA led to a better mean score for 16 of 17 instances.
The main difference between the two procedures is that SSF is a two-phase heuristic, whereas Relaxed ZGA fixes both capacity and battery constraints in a single pass.

\begin{table}[h]
\begin{tabularx}{\textwidth}{X|rrrr}
\hline
\\[-1em]
Method & BMS count & $\overline{mean\_scores}$ & BS count & $\overline{best\_scores}$ \\ 
\hline
VNS\_zga\_c:14\_p:3\_r:1 & 16 & 0.998 & 14 & 0.990 \\
VNS\_ssf\_c:14\_p:3\_r:1 & 1 & 1.002 & 6 & 0.993 \\
\hline
\end{tabularx}
\caption{VNS evolution: Replacement of the repair procedure}\label{tab:repl_repair}
\end{table}

\begin{table}[h]
\begin{tabularx}{\textwidth}{X|rrrr}
\hline
\\[-1em]
Method & BMS count & $\overline{mean\_scores}$ & BS count & $\overline{best\_scores}$ \\ 
\hline
VNS\_zga\_c:14\_ls:110\_p:3\_r:1 & 7 & 0.995 & 8 & 0.987 \\
VNS\_zga\_c:14\_ls:011\_p:3\_r:1 & 6 & 0.995 & 11 & 0.987 \\
VNS\_zga\_c:14\_ls:101\_p:3\_r:1 & 6 & 0.995 & 9 & 0.989 \\
VNS\_zga\_c:14\_ls:111\_p:3\_r:1 & 6 & 0.995 & 6 & 0.989 \\
VNS\_zga\_c:14\_ls:001\_p:3\_r:1 & 6 & 0.995 & 6 & 0.988 \\
VNS\_zga\_c:14\_ls:100\_p:3\_r:1 & 5 & 0.995 & 7 & 0.987 \\
VNS\_zga\_c:14\_ls:010\_p:3\_r:1 & 3 & 0.997 & 6 & 0.989 \\
VNS\_zga\_c:14\_ls:000\_p:3\_r:1 & 0 & 0.998 & 3 & 0.990 \\
\hline
\end{tabularx}
\caption{VNS evolution: Adding problem-specific operators}\label{tab:adding_opers}
\vspace{-1.5em}
\end{table}

\subsubsection{Adding problem-specific operators}\label{sec:adding_opers}
The last significant extension of the original GRASP is adding problem-specific operators: AFS-realloc-1 and AFS-realloc-more.
As the first operator addresses only subtours with exactly one AFS and the second only subtour with more than one, a third operator called AFS-realloc-all was considered in the tuning.
The AFS-realloc-all is only a sequential aggregation of the other two - the AFS-realloc-1 is run first,
and is followed by the AFS-realloc-more if it does not yield any improvement.
In Table~\ref{tab:adding_opers}, all possible combinations of these three operators were tested, which is encoded by the ls parameter. 
Each bit of the parameter encodes whether the corresponding operator was used in the local search.
For example, ls:101 means that AFS-realloc-1 was used, AFS-realloc-more was not used, and AFS-realloc-all was used.
It can be seen that the previous setup without any problem-specific operators is the worst in the experiment.
Most of the other variants give somewhat similar performance.
Based on the results, the variant ls:110 was selected.
This setup omits the combined AFS-realloc-all operator.

\subsubsection{Parameters tuning}\label{sec:param_tuning}

Finally, the VNS parameters $r$ (restart ratio) and $p$ (perturbation strength) were tuned in an exhaustive manner, by testing all combinations in the following ranges: $p\in\{1, 2, ..., 10\}$, $r\in\{0.05, 1, 1.5, ..., 10\}$. 
The parameters turned out to have a moderate effect on the VNS performance.
The best setup obtained is compared to the previous best in Table~\ref{tab:param_tuning}.
Note that the final VNS method's average performance is better than the best performance of the initial GRASP method.

\begin{table}[h]
\begin{tabular}{l|>{\raggedleft}p{0.13\textwidth}>{\raggedleft}p{0.14\textwidth}>{\raggedleft}p{0.13\textwidth}>{\raggedleft\arraybackslash}p{0.12\textwidth}}
\hline
\\[-1em]
Method & BMS count & $\overline{mean\_scores}$ & BS count & $\overline{best\_scores}$ \\ 
\hline
VNS\_zga\_c:14\_ls:110\_p:2\_r:0.35 & 11 & 0.995 & 14 & 0.988 \\
VNS\_zga\_c:14\_ls:110\_p:3\_r:1 & 9 & 0.995 & 9 & 0.989 \\

\hline
\end{tabular}
\caption{VNS evolution: Parameters tuning}\label{tab:param_tuning}
\vspace{-1.5em}
\end{table}

\subsection{Competition results}\label{sec:comp_results}

This section provides detailed results of the top three methods submitted to CEC-12: the VNS described in this paper, Simulated Annealing (SA), and Genetic Algorithm (GA).
Implementations of the latter two are not publicly available.
The results of the competition presented were generated on an identical machine by the organizers and are publicly available at~\textcite{competition}.
Apart from these, the Sequence-Based Hyperheuristic and Max-Min Ant System were also submitted.
The submitted VNS is configured according to the tuning process described in Section~\ref{sec:grasp_to_vns}.

\begin{table}[hb]
\begin{tabular}{l|rrr|rrr|rrr}
\hline
& \multicolumn{3}{c|}{VNS}          & \multicolumn{3}{c|}{SA}         & \multicolumn{3}{c}{GA} \\ 
Instance & min & mean & stdev & min & mean & stdev & min & mean & stdev \\ \hline
E-n22-k4    & \textit{\textbf{\textbf{384.}67}}    & \textit{\textbf{384.67}}    & 0.00   & \textit{\textbf{384.67}}    & \textit{\textbf{384.67}}    & 0.00   & \textit{\textbf{384.67}}    & \textit{\textbf{384.67}}    & 0.00   \\
E-n23-k3    & \textit{\textbf{571.94}}    & \textit{\textbf{571.94}}    & 0.00   & \textit{\textbf{571.94}}    & \textit{\textbf{571.94}}    & 0.00   & \textit{\textbf{571.94}}    & \textit{\textbf{571.94}}    & 0.00   \\
E-n30-k3    & \textit{\textbf{509.47}}    & \textit{\textbf{509.47}}    & 0.00   & \textit{\textbf{509.47}}    & \textit{\textbf{509.47}}    & 0.00   & \textit{\textbf{509.47}}    & \textit{\textbf{509.47}}    & 0.00   \\
E-n33-k4    & \textit{\textbf{840.14}}    & 840.43    & 1.18   & 840.57    & 854.07    & 12.80  & 844.25    & 845.62    & 0.92   \\
E-n51-k5    & \textit{\textbf{529.90}}    & 543.26    & 3.52   & 533.66    & 533.66    & 0.00   & \textit{\textbf{529.90}}    & 542.08    & 8.57   \\
E-n76-k7    & \textit{\textbf{692.64}}    & 697.89    & 3.09   & 701.03    & 712.17    & 5.78   & 697.27    & 717.30    & 9.58   \\
E-n101-k8   & \textit{839.29}    & 853.34    & 4.73   & 845.84    & 852.48    & 3.44   & 852.69    & 872.69    & 9.58   \\
X-n143-k7   & \textit{16028.05}  & 16459.31  & 242.59 & 16610.37  & 17188.90  & 170.44 & 16488.60  & 16911.50  & 282.30 \\
X-n214-k11  & \textit{11323.56}  & 11482.20  & 76.14  & 11404.44  & 11680.35  & 116.47 & 11762.07  & 12007.06  & 156.69 \\
X-n352-k40  & \textit{27064.88}  & 27217.77  & 86.20  & 27222.96  & 27498.03  & 155.62 & 28008.09  & 28336.07  & 205.29 \\
X-n459-k26  & \textit{25370.80}  & 25582.27  & 106.89 & 25809.47  & 26038.65  & 157.97 & 26048.21  & 26345.12  & 185.14 \\
X-n573-k30  & 52181.50  & 52548.11  & 278.85 & \textit{51929.24}  & 52793.66  & 577.24 & 54189.62  & 55327.62  & 548.05 \\
X-n685-k75  & \textit{71345.40}  & 71770.57  & 197.08 & 72549.90  & 73124.98  & 320.07 & 73925.56  & 74508.03  & 409.43 \\
X-n749-k98  & \textit{81002.01}  & 81327.39  & 176.19 & 81392.78  & 81848.13  & 275.26 & 84034.73  & 84759.79  & 376.10 \\
X-n819-k171 & \textit{164289.95} & 164926.41 & 318.62 & 165069.77 & 165895.78 & 403.70 & 170965.68 & 172410.12 & 568.58 \\
X-n916-k207 & \textit{341649.91} & 342460.70 & 510.66 & 342796.88 & 343533.85 & 556.98 & 357391.57 & 360269.94 & 229.19 \\
X-n1001-k43 & \textit{77476.36}  & 77920.52  & 234.73 & 78053.86  & NA        & 306.27 & 78832.90  & 79163.34  & NA     \\ \hline
\end{tabular}
\caption{Official results of the IEEE WCCI 2020 Competition on EVRP}\label{tbl:results_abs}
\vspace{-1em}
\end{table}

\begin{table}[h]
\begin{tabular}{l|rr|rr|rr}
\hline
& \multicolumn{2}{c|}{VNS}          & \multicolumn{2}{c|}{SA}         & \multicolumn{2}{c}{GA} \\
Instance        & \shortstack{min \\ gap [\%]} & \shortstack{mean \\ gap [\%]} & \shortstack{min \\ gap [\%]} & \shortstack{mean \\ gap [\%]} & \shortstack{min \\ gap [\%]} & \shortstack{mean \\ gap [\%]} \\ \hline
E-n22-k4	    & \textit{\textbf{0.00}}  & \textit{\textbf{0.00}}	& \textit{\textbf{0.00}}  & \textit{\textbf{0.00}}	& \textit{\textbf{0.00}}  & \textit{\textbf{0.00}} \\
E-n23-k3	    & \textit{\textbf{0.00}}  & \textit{\textbf{0.00}}	& \textit{\textbf{0.00}}  & \textit{\textbf{0.00}}	& \textit{\textbf{0.00}}  & \textit{\textbf{0.00}} \\
E-n30-k3	    & \textit{\textbf{0.00}}  & \textit{\textbf{0.00}}	& \textit{\textbf{0.00}}  & \textit{\textbf{0.00}}	& \textit{\textbf{0.00}}  & \textit{\textbf{0.00}} \\
E-n33-k4	    & \textit{\textbf{0.00}}  & 0.03	& 0.05  & 1.66	& 0.49  & 0.65 \\
E-n51-k5	    & \textit{\textbf{0.00}}  & 2.52	& 0.71  & 0.71	& \textit{\textbf{0.00}}  & 2.30 \\
E-n76-k7	    & \textit{\textbf{0.00}}  & 0.76	& 1.21  & 2.82	& 0.67  & 3.56 \\
E-n101-k8	    & \textit{0.61}  & 2.29	& 1.39  & 2.19	& 2.21  & 4.61 \\
X-n143-k7	    & \textit{0.95}  & 3.66	& 4.62  & 8.26	& 3.85  & 6.51 \\
X-n214-k11	    & \textit{1.89}  & 3.32	& 2.62  & 5.10	& 5.84  & 8.04 \\
X-n351-k40	    & \textit{2.22}  & 2.79	& 2.81  & 3.85	& 5.78  & 7.02 \\
X-n459-k26	    & \textit{2.87}  & 3.73	& 4.65  & 5.58	& 5.62  & 6.82 \\
X-n573-k30	    & 1.58  & 2.30	& \textit{1.09}  & 2.77	& 5.49  & 7.71 \\
X-n685-k75	    & \textit{2.43}  & 3.04	& 4.16  & 4.99	& 6.13  & 6.97 \\
X-n749-k98	    & \textit{2.16}  & 2.57	& 2.66  & 3.23	& 5.99  & 6.90 \\
X-n819-k171	    & \textit{1.82}  & 2.21	& 2.30  & 2.81	& 5.95  & 6.85 \\
X-n916-k207	    & \textit{1.90}  & 2.14	& 2.24  & 2.46	& 6.59  & 7.45 \\
X-n1001-k43	    & \textit{3.22}  & 3.82	& 3.99  & NA	& 5.03  & 5.47 \\
\hline
\end{tabular}
\caption{Comparison of the individual methods relative to BKSs}\label{tbl:results_rel}
\vspace{-1em}
\end{table}

Table~\ref{tbl:results_abs} provides the best scores obtained by each of the top three methods, together with the mean and standard deviation values.
The best scores from the competition are written in italics.
For some instances, multiple methods reached the best known score (BKS).
Scores of such are written in bold.
Values marked NA are not provided in the competition results.
It can be seen that the VNS produced the best score for 16 of 17 instances and the BMS for 15 of 17 instances. 
In the three smallest instances, the BKS was obtained by each of the top three methods in every run. 
The SA method yields better average performance in two instances: E-n51-k5 and E-n101-k8; however, it does not reach the best score for these.
Overall, the VNS consistently outperforms the other two top methods in a fair comparison with an identical stop condition. 

Table~\ref{tbl:results_rel} shows the same data relative to BKSs.
The gap values are calculated as $gap=100(\frac{score}{BKS}-1)[\%]$, where $score$ is either the minimum or mean value.
For the 8 smallest instances, the VNS consistently finds a best solution within $1\%$ from the BKS and generally does not exceed $4\%$ in both best and average performance.
For the remaining two methods, the best solution is always within $5\%$ for the SA and $7\%$ for the GA.
In terms of average performance, both methods exceed the $8\%$ mean gap in a single instance.
SA performs significantly better than GA on the larger X instances.

\begin{table}[h]
\begin{tabular}{l|rrrrr|rrrrr}
\hline
& \multicolumn{5}{c|}{VNS}          & \multicolumn{5}{c}{BACO}  \\ 
Instance	&	min	&	gap	[\%] &	mean	&	gap	[\%] &	stdev	&	min	&	gap	[\%] &	mean	&	gap	[\%] &	stdev	\\ \hline
E-n22-k4	&	\textbf{384.67}	&	0.00	&	\textbf{384.67}	&	0.00	&	0.00	&	\textbf{384.67}	&	0.00	&	\textbf{384.67}	&	0.00	&	0.00	\\
E-n23-k3	&	\textbf{571.94}	&	0.00	&	\textbf{571.94}	&	0.00	&	0.00	&	\textbf{571.94}	&	0.00	&	\textbf{571.94}	&	0.00	&	0.00	\\
E-n30-k3	&	\textbf{509.47}	&	0.00	&	\textbf{509.47}	&	0.00	&	0.00	&	\textbf{509.47}	&	0.00	&	\textbf{509.47}	&	0.00	&	0.00	\\
E-n33-k4	&	\textbf{840.14}	&	0.00	&	\textbf{840.14}	&	0.00	&	0.00	&	840.57	&	0.05	&	842.30	&	0.26	&	1.42	\\
E-n51-k5	&	\textbf{529.90}	&	0.00	&	\textbf{529.90}	&	0.00	&	0.00	&	\textbf{529.90}	&	0.00	&	\textbf{529.90}	&	0.00	&	0.00	\\
E-n76-k7	&	\textbf{692.64}	&	0.00	&	\textbf{692.64}	&	0.00	&	0.00	&	\textbf{692.64}	&	0.00	&	692.85	&	0.03	&	0.81	\\
E-n101-k8	&	\textbf{834.22}	&	0.00	&	\textbf{834.73}	&	0.06	&	0.69	&	840.25	&	0.72	&	845.95	&	1.41	&	4.58	\\
X-n143-k7	&	\textbf{15877.50}	&	0.00	&	15888.37	&	0.07	&	5.97	&	15901.23	&	0.15	&	16031.46	&	0.97	&	262.47	\\
X-n214-k11	&	\textbf{11113.20}	&	0.00	&	11144.77	&	0.28	&	18.24	&	11133.14	&	0.18	&	11219.70	&	0.96	&	46.25	\\
X-n351-k40	&	26552.20	&	0.28	&	26625.39	&	0.56	&	39.38	&	\textbf{26478.34}	&	0.00	&	26593.18	&	0.43	&	72.86	\\
X-n459-k26	&	\textbf{24663.00}	&	0.00	&	24773.50	&	0.45	&	41.66	&	24763.93	&	0.41	&	24916.60	&	1.03	&	94.08	\\
X-n573-k30	&	\textbf{51368.40}	&	0.00	&	51485.92	&	0.23	&	47.70	&	53822.87	&	4.78	&	54567.15	&	6.23	&	231.05	\\
X-n685-k75	&	\textbf{69652.50}	&	0.00	&	69845.30	&	0.28	&	91.86	&	70834.88	&	1.70	&	71440.57	&	2.57	&	281.78	\\
X-n749-k98	&	\textbf{79285.80}	&	0.00	&	79565.06	&	0.35	&	121.90	&	80299.76	&	1.28	&	80694.54	&	1.78	&	223.91	\\
X-n819-k171	&	\textbf{161361.00}	&	0.00	&	161765.06	&	0.25	&	203.43	&	164720.80	&	2.08	&	165565.79	&	2.61	&	401.02	\\
X-n916-k207	&	\textbf{335282.00}	&	0.00	&	336076.81	&	0.24	&	427.48	&	342993.01	&	2.30	&	344999.95	&	2.90	&	905.72	\\
X-n1001-k43	&	\textbf{75055.90}	&	0.00	&	75348.39	&	0.39	&	168.81	&	76297.09	&	1.65	&	77434.33	&	3.17	&	719.86	\\
\hline
\end{tabular}
\caption{Comparison with BACO}\label{tab:BACO}
\vspace{-1.5em}
\end{table}

\subsection{Comparison with BACO}\label{sec:BACO}

This section provides a separate comparison with the Bilevel Ant Colony Optimization (BACO) algorithm proposed by~\textcite{Jia2021} after the competition.
As explained in Section~\ref{sec:results},~\textcite{Jia2021} used a different stop condition from the competition.
The results presented in~\textcite{Jia2021} are given in Table~\ref{tab:BACO} together with our VNS results.
Both methods are given the same time budget, defined in Equation~\ref{eq:t_max_stop}.
The BACO implementation is not available for direct comparison on an identical machine.
In order to provide as a fair comparison as possible, the time budget given to the VNS is adjusted proportionately to the difference in the performance of the CPUs used in~\textcite{Jia2021} and in this paper (see Equation~\ref{eq:vns_stop}).
Again, the BKS are written in bold.
Table~\ref{tab:BACO} also provides the gap values from the BKS.

\begin{figure}[h]
\centering{\includegraphics[width=0.7\textwidth]{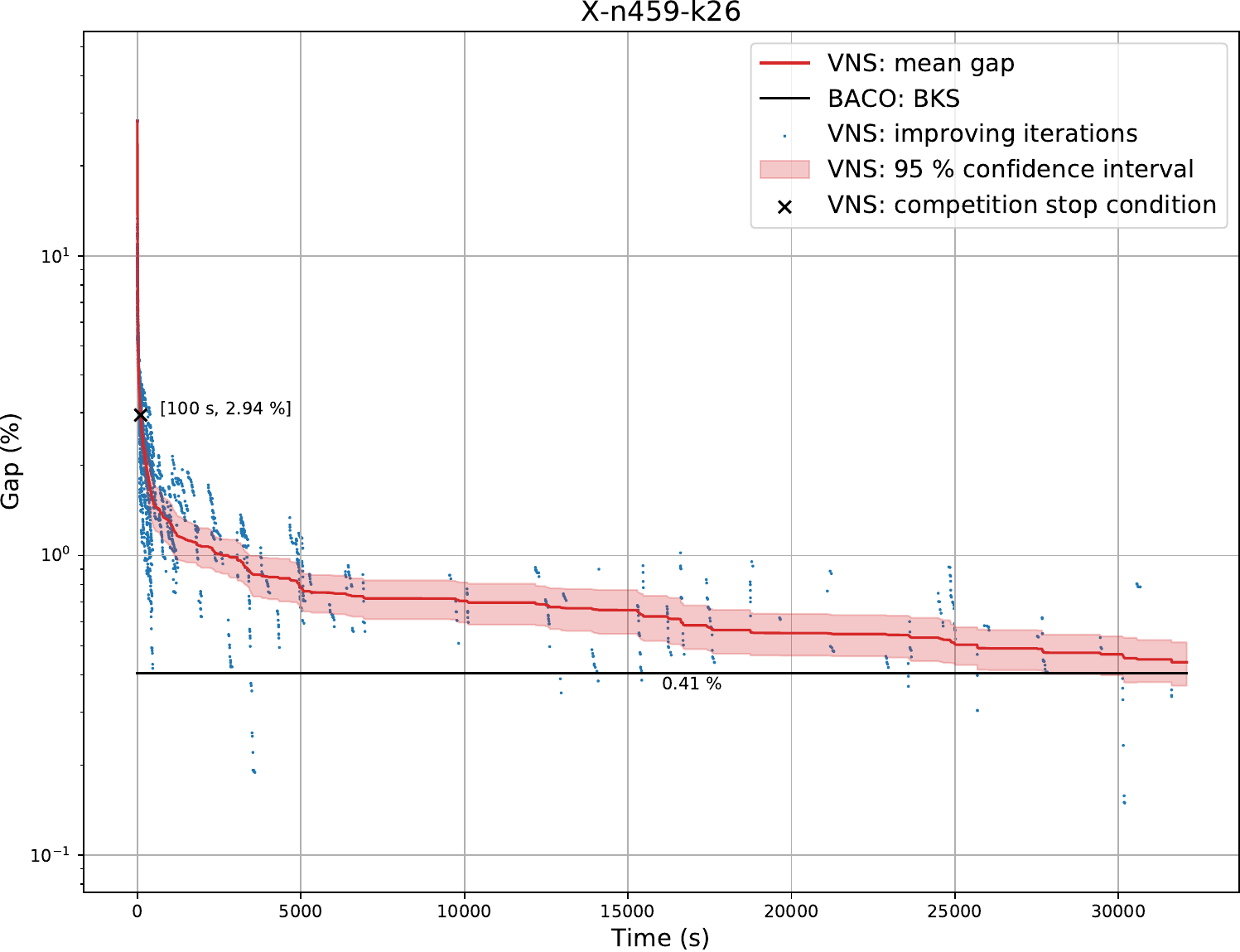}}
   \caption{Progress of solution quality - instance X-n459-k26}
   \label{fig:progress_n459}
   \vspace{-1.5em}
\end{figure}

\begin{figure}[h]
\centering{\includegraphics[width=0.7\textwidth]{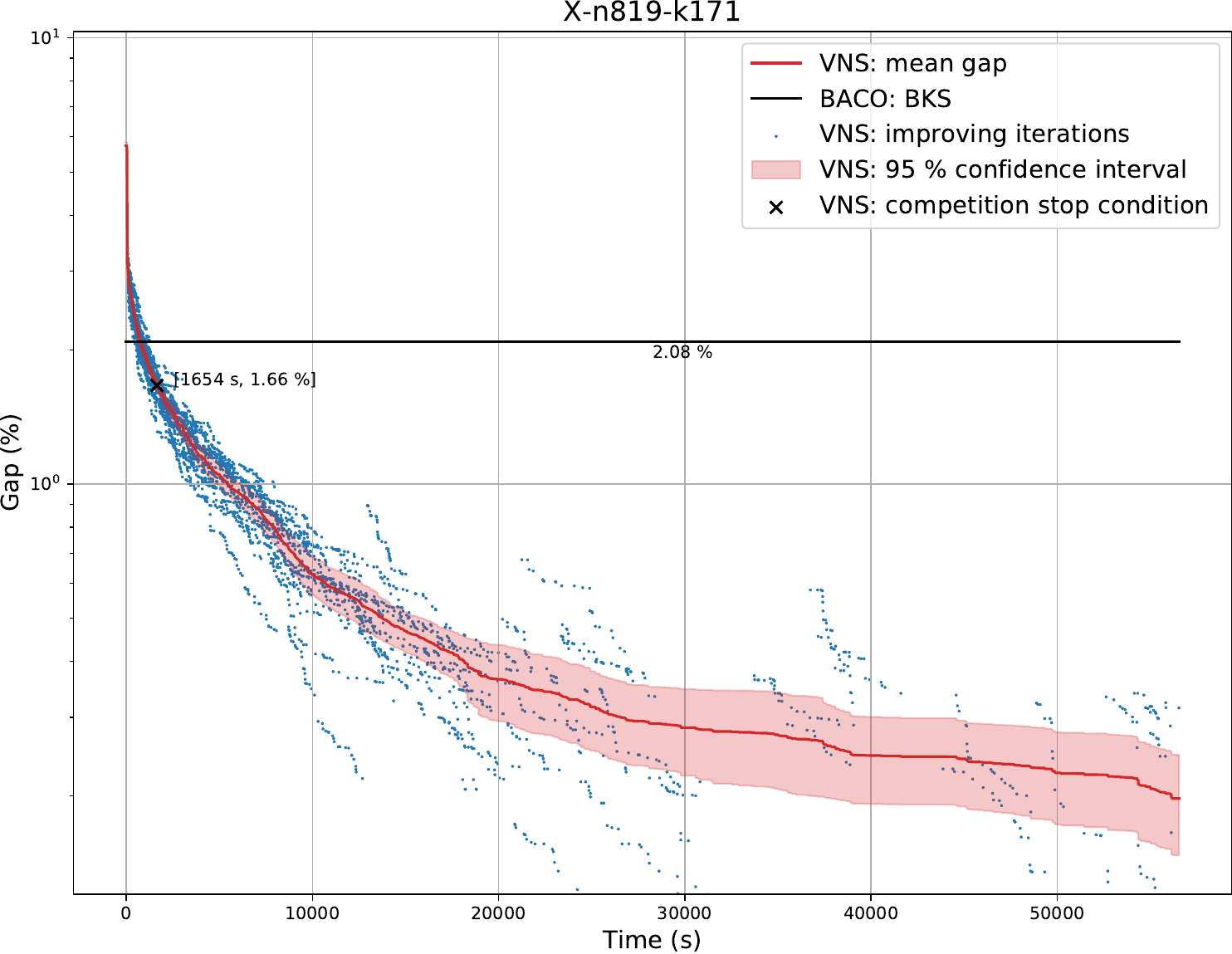}}
   \caption{Progress of solution quality - instance X-n819-k171}
   \label{fig:progress_n819}
   \vspace{-1.5em}
\end{figure}

The VNS found the BKS in 16 of 17 instances, while 10 of these were newly found in this experiment.
The BACO then found the BKS on 6 instances, 5 of which are smaller E instances, where the BKS is reached in almost every run by both methods.
As for the mean score, the trend is very similar. 
Therefore, the BACO seems to yield competitive performance on small and medium-sized instances but does not scale as well as the VNS.
The results presented in Table~\ref{tab:BACO} show that after adopting the same stop condition, the VNS performance on the competition dataset is superior to that of BACO.
The progress in solution quality of the VNS metaheuristic in two instances is illustrated in Figure~\ref{fig:progress_n459} and Figure~\ref{fig:progress_n819}.
Both figures show the VNS mean gap, $95\%$ confidence interval, and the mean gap value corresponding to the competition stop condition given by the number of fitness evaluations (black cross).
Although its exact value depends on the machine used, it is evident that the competition stop condition is significantly stricter.
The mean gap values and confidence intervals presented were calculated from the fitness values of the best overall solution $T^{**}$, which were recorded every second.
The blue dots in both figures show these values, but only after improving VNS iterations, i.e. whenever $T^{**}$ changes.
In addition, the best score obtained by BACO, as reported in~\textcite{Jia2021}, is shown (BACO: BKS, black line).
In case of the larger instance X-n819-k171, the BACO best solution is consistently reached by the VNS after using approximately $2\%$ of the time budget and the final VNS mean gap is lower by almost $2\%$ than the BACO best solution.
Given the BACO gap values shown in Table~\ref{tab:BACO}, this trend is likely to be similar in the case of all instances larger than X-n459-k26.

\section{Conclusions}\label{sec:conclusions}

EVRP is a highly relevant problem that naturally arises due to the current trends in the transport industry.
This paper presents a VNS method, which won the CEC-12 Competition on Electric Vehicle Routing Problem, assigned within IEEE WCCI 2020.
The method is compiled from various previously known algorithms, and some of them had to be tailored for the EVRP.
The paper documents the evolution of the method from our previous GRASP method; thus, individual components' performance improvement can be determined.
It is also shown which components were tested but turned out not to be contributive.
Thanks to the competition, a fair comparison with the remaining four competition entries is guaranteed.
Such a portfolio of compared methods is otherwise rare due to the large variety in EVRP formulations.
The method is also compared with a BACO algorithm, proposed after the competition, and outperforms it as well, while finding new BKSs to a large portion of the competition dataset.
Thus, the developed method is state of the art on a fundamental variant of the EVRP.

In future work, it will be researched to what extent this method will be competitive when extended to more complex formulations in a way that preserves the current implementation as much as possible.
In these formulations, the solution representation is often the same, but they usually introduce additional constraints and optimization criteria.
It will be studied which formulations can be efficiently solved by adjustments such as fitness function redefinition, solution invalidity penalization, or adding black-box validity checking.

\section*{Declarations}



\noindent \textbf{Data availability statement} \\
The data that support the findings of this study, including source codes and the competition dataset, are openly available in a persistent GitHub repository at \hyperlink{a}{https://github.com/wolledav/VNS-EVRP-2020}.

\printbibliography

\end{document}